\documentclass[lettersize,journal]{IEEEtran}

\usepackage{array}
\usepackage[caption=false,font=normalsize,labelfont=sf,textfont=sf]{subfig}
\usepackage{textcomp}
\usepackage{stfloats}
\usepackage{url}
\usepackage{verbatim}
\usepackage{cite}
\usepackage{amsmath,amssymb,amsfonts}
\usepackage{graphicx}
\usepackage{textcomp}
\usepackage{xcolor}
\usepackage[abs]{overpic}
\usepackage{algorithm}
\usepackage[end]{algpseudocode}
\usepackage{etoolbox}
\usepackage{subcaption}
\usepackage{hyperref}
\usepackage[inline]{enumitem}
\usepackage{flushend}

\def\BibTeX{{\rm B\kern-.05em{\sc i\kern-.025em b}\kern-.08em
    T\kern-.1667em\lower.7ex\hbox{E}\kern-.125emX}}
\usepackage{balance}

\begin{document}
\title{RoboCar: A Rapidly Deployable Open-Source Platform for Autonomous Driving Research}

\author{Mehdi~Testouri,
        Gamal~Elghazaly,~\IEEEmembership{Member,~IEEE,}
        and~Raphael~Frank,~\IEEEmembership{Senior Member,~IEEE}
\thanks{All authors are within 
SnT - Interdisciplinary Centre for Security, Reliability and Trust, University of Luxembourg. 
29 Avenue John F. Kennedy, 1855 Luxembourg, 
e-mail: \{ mehdi.testouri, gamal.elghazaly, raphael.frank\}@uni.lu}
}

\maketitle

\begin{abstract}
This paper introduces RoboCar, an open-source research platform for autonomous driving developed at the University of Luxembourg. RoboCar provides a modular, cost-effective framework for the development of experimental Autonomous Driving Systems (ADS), utilizing the 2018 KIA Soul EV. The platform integrates a robust hardware and software architecture that aligns with the vehicle’s existing systems, minimizing the need for extensive modifications. It supports various autonomous driving functions and has undergone real-world testing on public roads in Luxembourg City. This paper outlines the platform's architecture, integration challenges, and initial test results, offering insights into its application in advancing autonomous driving research. RoboCar is available to anyone at \href{https://github.com/sntubix/robocar}{\textcolor{magenta}{https://github.com/sntubix/robocar}} and is released under an open-source MIT license.
\end{abstract}


\section{Introduction}
\IEEEPARstart{A}{utonomous} Vehicles (AVs) represent a transformative advancement in transportation, offering increased safety and efficiency. Despite substantial progress over the last decade, developing Autonomous Driving Systems (ADS) involves intricate challenges across technological, regulatory, and ethical realms.

\begin{figure}
    \centering
    \includegraphics[scale=0.44]{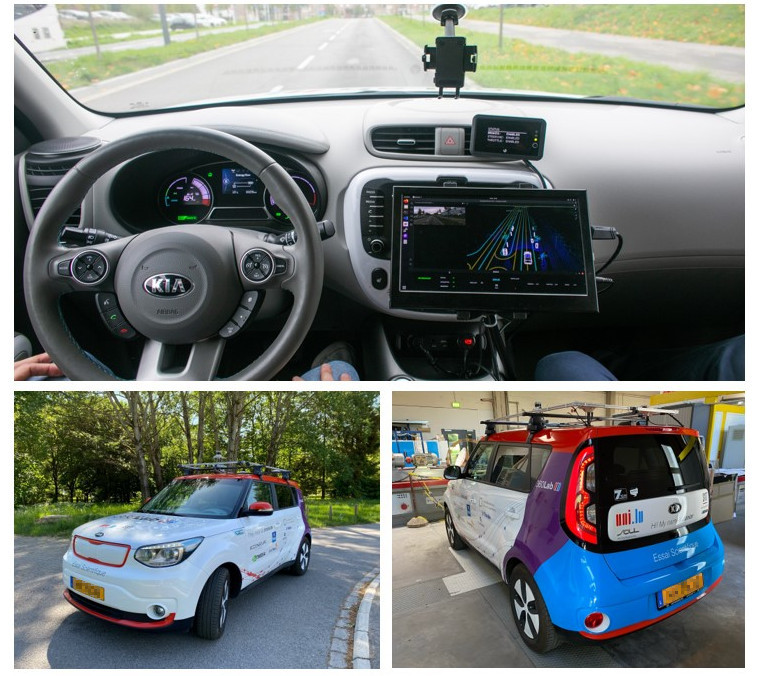}
    \caption{\footnotesize The experimental vehicle based on a 2018 KIA Soul EV retrofitted with hardware and sensors enabling self-driving operation.}
    \label{fig:junior}
\end{figure}

This research introduces RoboCar, an open-source research platform developed at the University of Luxembourg. RoboCar provides a modular, cost-effective autonomous driving solution tailored for, but not limited to, the 2018 KIA Soul EV (see Fig. \ref{fig:junior}), enabling advanced research and practical applications in autonomous driving technologies.

The RoboCar initiative aims to facilitate and accelerate the development of experimental ADS by offering an easy-to-setup framework that includes all the relevant modules to test and operate an ADS. As middleware, we rely on ROS2 allowing for greater compatibility and fast integration. The aim is to propel academic research and collaborative innovation. As an open-source project, it contributes to the global efforts towards next-generation autonomous systems. In this paper, we describe RoboCar's architecture, the hardware and software platform, including various integration challenges we faced. Finally, we will present the results of an experimental trial conducted on public roads in Luxembourg City.

The remainder of the paper is organized as follows: Section II reviews related works. Section III and IV provide a detailed description of the Hardware and Software platforms, respectively. Section V presents the findings of the experimental trial before discussing the scope and limitations of the platform in Section VI. We conclude the paper in Section VII.

\section{Related work}

Autonomous vehicle research has come a long way since Stanley, the autonomous vehicle of Stanford University, won the DARPA Grand Challenge in 2005~\cite{darpa}. Since the early days, universities and research centers have been important drivers in designing and improving ADS. A plethora of research articles have been published over the last 15 years. A comprehensive survey capturing the important trends and emerging technologies has been published by Yurtsever et al.~\cite{survey20}. In addition to offering a compact overview of state-of-the-art architectures and technologies, it also lists the major open-source initiatives and how they compare. A similar work has been published by Badue et al. ~\cite{survey21}. This work also includes a description of the Intelligent Autonomous Robotics Automobile (IARA), a platform developed at the Federal University of Espírito Santo in Brazil. Unfortunately, this project has not been open-sourced.

The first attempts to launch an open-source project for autonomous driving began in 2015 by Kato et al. at Nagoya University in Japan. The project, named Autoware~\cite{autoware15gh}, introduced an open platform using commodity vehicles and sensors. The initial article presents algorithms, software libraries, and datasets required for scene recognition, path planning, and vehicle control \cite{autoware15}.

Another prominent open-source project is Baidu's Apollo\cite{apollo17gh}. The initial development started in 2017 with the release of Apollo 1.0. Over the years, the platform continued to evolve at an impressive pace, forming strategic partnerships with various tech and automotive companies. Today, the release version 8.0 specifies hardware and software components and includes a simulation suite.

A work that describes the challenges, pitfalls, and lessons learned from integrating and running the open-source Apollo driving stack into a research vehicle has been published by Kessler et al.~\cite{fortuna19}. They conclude that the integration of such a complex software platform with specific vehicle hardware is not straightforward but achievable with the right engineering expertise.

Another open-source project introduced by comma.ai is Openpilot~\cite{openpilot22, openpilot22gh}. Unlike Autoware and Apollo, which are full-stack autonomous driving platforms, Openpilot functions primarily as an L2 Advanced Driver Assistance System (ADAS). It integrates with a specific hardware module that employs a camera system and connects to a vehicle's onboard systems. This configuration enables compatibility with over 250 commercial car models. It implements functions including Automated Lane Centering, Adaptive Cruise Control, and Lane Change Assist.

A more recent initiative gaining increasing attention is the open-source project Aslan~\cite{aslangh}. It has been specifically designed for low-speed urban environments. It defines a complete self-driving software stack and includes a simulator supported by Gazebo~\cite{gazebo}. The project is a collaborative effort involving various stakeholders, including technology companies, universities, and public authorities.

Similarly, the European project PRYSTINE~\cite{nov22} used the same vehicle platform as RoboCar to enhance automotive architectures for automated driving by focusing on fail-operational systems. As the main focus of this project was to investigate fail-operational perception based on sensor fusion, it did not implement a full autonomous driving stack, nor has the implementation been open-sourced.

Finally, the AnnieWAY team, composed of researchers from the Karlsruhe Institute of Technology and the FZI Research Center for Information Technology in Germany, participated in the 2016 Grand Cooperative Driving Challenge and developed an automated vehicle based on a modified Mercedes-Benz S-Class platform, dubbed BerthaOne. They specified a hardware and software stack for perception, cooperation, and motion planning~\cite{8055618}. Unfortunately, the software stack has not been open-sourced.

In addition to the software and hardware platforms that can be utilized for experimental studies, open-source simulators have also seen a surge in popularity over recent years, as they eliminate the need to purchase expensive hardware. The most well-regarded among these include Carla\cite{carla17}, LG SVL~\cite{lgsvl20}, and AirSim~\cite{airsim17}. Each of these simulators implement the full stack of an autonomous vehicle and rely on a high-fidelity 3D engine to create realistic driving environments. However, LG SVL and AirSim have recently been discontinued, leaving Carla as the only full-stack option.

The research outlined in this paper is the result of more than five years of extensive work across various projects~\cite{8782528, junior19, testouri2022fastcycle, testouri2023towards}. It distinguishes itself from prior works in several key ways: (1) it introduces a software and hardware architecture specifically designed for deployment on a specific commercial vehicle; (2) the platform is cost-effective, making it well-suited for research and startup settings; (3) it is designed to integrate seamlessly with the vehicle's existing systems, removing the need to add or modify the embedded systems, and thus does not require specialized engineering skills as to speed up the integration.

\section{Hardware Platform}
\begin{figure}[!t]
    \centering
    \includegraphics[scale=0.39]{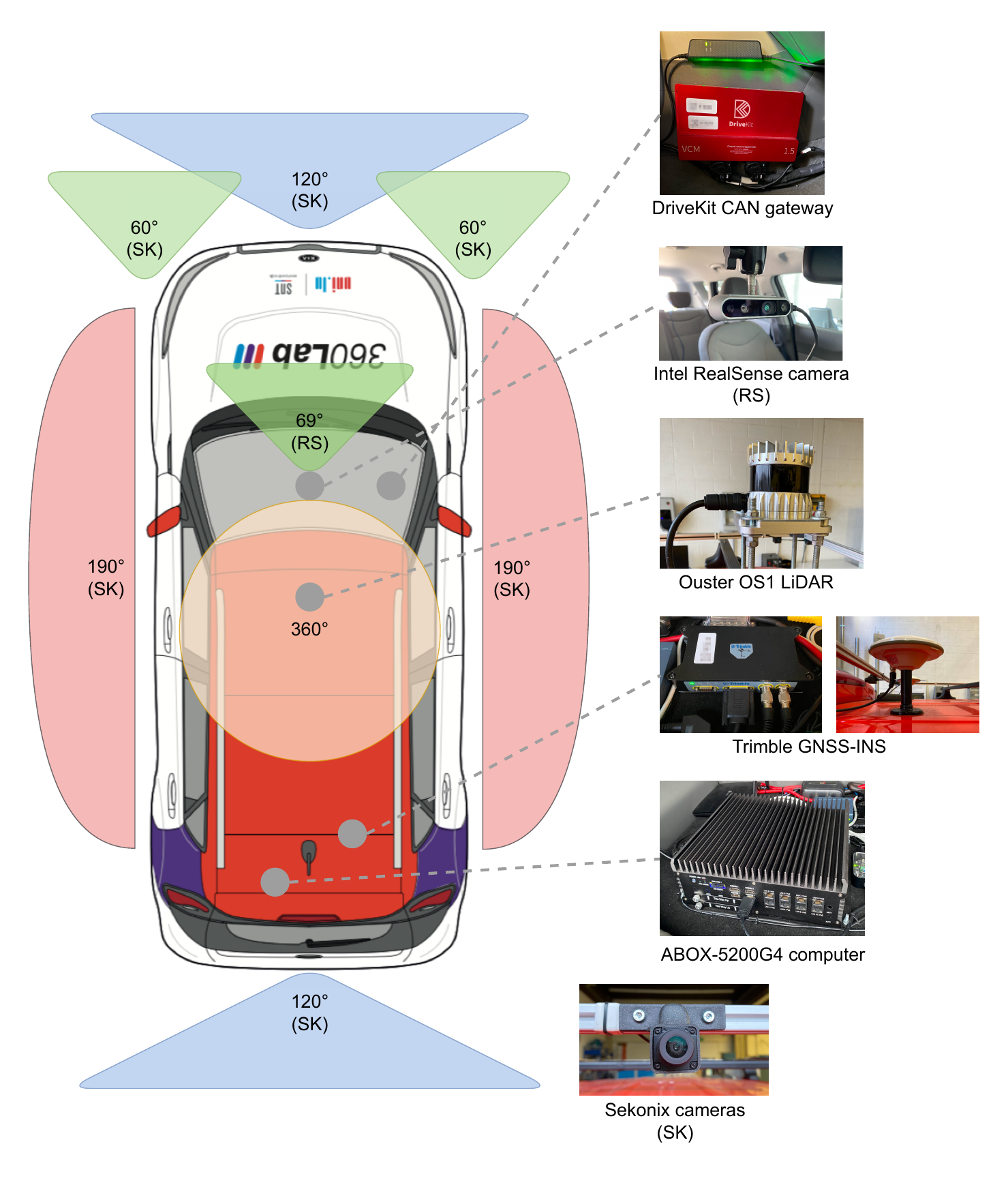}
    \caption{\footnotesize Summary of the cost-efficient hardware platform. The LiDAR provides a 360 degrees point cloud representation around the vehicle at a range of 100m while the cameras cover all viewpoints around the vehicle. The onboard GNSS-INS has centimeter level accuracy.}
    \label{fig:hardware}
\end{figure}

This section describes the hardware platform used to deploy and test RoboCar as an ADS. Many combinations of hardware can be envisioned for different objectives and requirements. The scope of this work being research, the components were carefully chosen as to provide best trade-off between performance and affordability thus making it sound for a broad range of research groups and innovation projects. A summary of the hardware platform is shown on Fig. \ref{fig:hardware}.

\subsection{Experimental vehicle}
\subsubsection{Vehicle type}
The test vehicle is a KIA Soul EV model 2018. This vehicle has been selected because it has built-in drive-by-wire capabilities, meaning that it can easily be controlled by sending actuation commands over the CAN bus. Being electric, the vehicle longitudinal control is also made easier, moreover, maintenance and operating cost are also kept low. The experimental vehicle was named \textit{Junior} \cite{junior19} and is shown on Fig. \ref{fig:junior}.

\subsubsection{Vehicle interface}
As the experimental vehicle is a regular commercial car, an interface providing an easy access to the CAN bus proved necessary. The DriveKit from the company Polysync\footnote{https://www.polysync.io} was used for this purpose. It is based on the Open Source Car Control (OSCC) project \cite{oscc}, which provides an interface to the control systems of the vehicle and allows to retrieve information such as steering wheel position or vehicle speed.

\subsection{Onboard equipment and sensors}
\subsubsection{Localization}
Localization techniques rely mainly on fusing different sources of information such as vehicle odometry, GNSS, IMU, and even map data \cite{chalvatzaras2022survey}. 
Essential to get a precise vehicle localization is to find a suitable way to fuse these data to estimate the vehicle pose \cite{shan2023survey}. For simplicity and reliability, localization in Robocar is based on an all-in-one solution relying on a high-precision INS namely the Trimble BX992 capable of providing a centimeter-level positioning and a precise heading thanks to two antennas mounted on the roof. The Trimble BX992-INS unit estimates vehicle pose using an extended kalman filter (EKF) fusing positional information from RTK-corrected GNSS position with data coming from an embedded IMU. The unit, which runs at 20Hz, also supplies INS full-navigation data including acceleration and speed characteristics of the vehicle.

\subsubsection{LiDAR}
The LiDAR is a crucial component of the perception as it allows to construct a 3D scene around the vehicle using a point cloud, which is then used to detect and track various relevant objects in proximity of the vehicle \cite{pointpillars2018} \cite{voxelnext2023}. LiDARs are also crucial components to build maps \cite{elghazaly2023high} and help to refine the vehicle position, especially in GNSS-denied areas \cite{massa2020lidar}. These sensors can be very expensive, as such, a compromise was found in the Ouster OS1 which provides decent performance at a fair price. This 20Hz LiDAR has a 360 degrees field of view with a range of 100m and a vertical resolution of 64 channels.

\subsubsection{Cameras}
Cameras complement the LiDAR perception and are also used for object detection. The experimental vehicle was equipped with six Sekonix GSML cameras providing a resolution of up to 1928x1208 at 30fps and horizontal FOVs of 60, 120 and 190 degrees. The cameras were mounted on the roof covering the front, rear and both sides, the purpose of this setup being to cover all viewpoints around the vehicle but it can prove to be expensive. An Intel RealSense D435 mounted on the windshield inside the cabin was also added and has an RGB resolution of 1920×1080 at 30fps and a horizontal FOV of 69 degrees in addition to a depth resolution up to 1280×720 at 90fps. The RealSense provides an alternative configuration for the camera perception where it relies on a single camera system similar to comma.ai devices \cite{openpilot22}\cite{openpilot22gh}. This setup is simpler and more cost-effective but potentially limiting depending on the application. For more complex perception scenarios, a wider-angle camera than the RealSense would be best suited. If used in conjunction with the Sekonix cameras, the RealSense camera can be used instead for in-cabin monitoring.

\subsubsection{Onboard computing}
Being specifically designed for such purposes, the computer running the ADS is a Sintrones ABOX-5200G4 equipped with an Intel Core i7-8700T, a GTX 1060 and 32GB of RAM. This unit provides a good balance of power efficiency and cooling while remaining powerful enough for resource intensive software and moderate AI duties.

\subsubsection{Connectivity}
A TP-Link Archer MR600 router is used to provide the experimental vehicle with internet access, but also to build an internal network that allows the different computers and sensors to communicate together.

\section{Software Platform}

\begin{figure}[!t]
    \centering
    \includegraphics[scale=0.55, clip, trim=1.0cm 18.5cm 0.0cm 1.0cm]{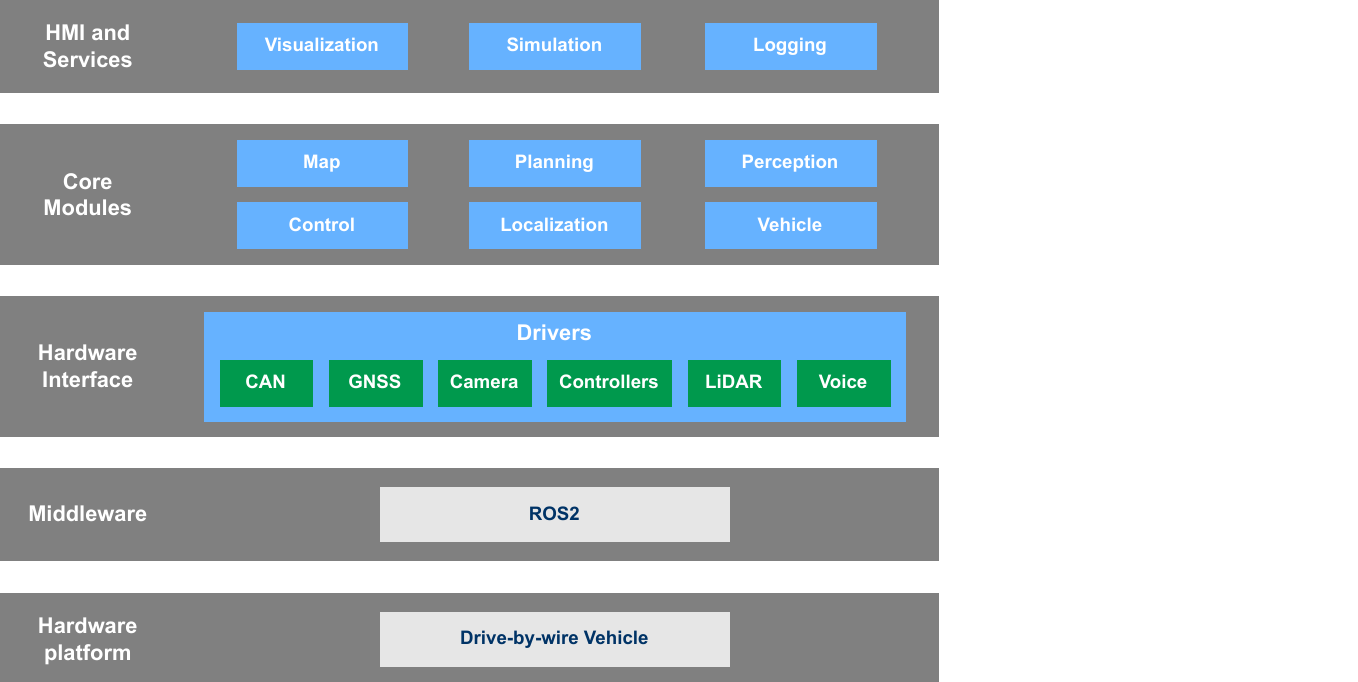}
    \caption{\footnotesize Architecture overview of RoboCar with the different modules represented in blue. The components constituting the drivers module are detailed in green. The middleware is based on ROS2 for greater compatibility and integration.}
    \label{fig:robocar}
\end{figure}

This section describes the in-house modular and easy to extend software developed as a research platform for connected and automated driving. Although initially using a custom middleware, the software has recently been ported to ROS2, allowing for greater compatibility and fast integration. The code is written in C++ and Python.

\subsection{Architecture}
Similar to Apollo \cite{apollo17gh}, RoboCar is based on a modular architecture with modules built as ROS2 packages and containing components. A component is an individual software unit responsible for a specific task and each component is only aware of itself, thus relying on the middleware to exchange messages. A component can also periodically execute a function, using its own thread. RoboCar is multithreaded but not multiprocess, meaning that all messages are exchanged in an intra-process manner and the components constitute the ROS2 nodes. This increases the speed of message transfer, especially for larger messages such as images and point clouds. The modules can be grouped into three main categories, namely (1) the human-machine interface and services, (2) the core modules and (3) the hardware interface comprising of the drivers. Those modules are built upon the middleware, itself based on ROS2. A summary of the RoboCar software is shown in Fig. \ref{fig:robocar}. The following sections will describe the different modules and their components in more detail.

\subsection{Middleware}
The middleware is the core library used to define the components and provide them with a configuration manager allowing to define parameters and to enable or disable components in a flexible manner. The library is header-only and easy to integrate. The middleware module was initially developed as part of the FastCycle project \cite{testouri2022fastcycle} which aimed at providing a high-performance intra-process middleware. Although FastCycle was faster than ROS1, the performance were not significantly better than ROS2, as a consequence, we added support for ROS2 as it provides greater flexibility and is used by a large community.

\subsection{Hardware interface}
The hardware interface allows for the rest of the software modules to be abstracted from the specific vehicle, devices and sensors employed. In RoboCar, this abstraction has been implemented in the Drivers module.
\subsubsection{Drivers}
The drivers module contains all components implementing drivers for the devices and sensors used in RoboCar such as the LiDAR, cameras, GNSS, etc. The drivers module should be the only one requiring changes or additions in order to run RoboCar on another hardware platform and vehicle. Core driver components are included for the CAN (DriveKit), the GNSS, the cameras and the LiDAR. In addition, RoboCar has a driver for joystick controllers allowing to drive the car using a game-pad or a similar device and a primitive driver for a voice command integration.

\subsection{Core modules}
The core modules comprise six modules essential for autonomous driving operation that do not rely on a specific hardware platform. Those modules are:
\subsubsection{Vehicle}
This module acts as an interface between RoboCar and the experimental vehicle. A commander component continuously checks the validity of sensor data such as timestamp and accuracy, as well as other safety parameters such as steering speed, and decides accordingly of autonomous driving engagement or disengagement. If the data coming from sensors are deemed unreliable or if the vehicle is performing maneuvers outside safety limits, the system may issue warnings or completely disengage the autonomous driving mode. The commander will also send actuation commands coming from the control after a series of checks and processing.

\subsubsection{Localization}
The localization is an essential part of the autonomous driving software as it provides the vehicle with an accurate position and orientation within a reference frame upon which the remaining core components in the autonomous driving stack rely on to function properly. 
Localization in the current release of RoboCar is based on the fusion of IMU data and RTK-corrected GNSS using extended EKF embedded in the Trimble BX992-INS. This compact solution provides an accuracy of 0.05m in horizontal positioning, 0.03m in vertical positioning, and 0.09 deg. in heading in RTK mode. In GNSS denied areas, as in case of tunnels, the Trimble BX992-INS estimated pose degrades to 0.30m horizontal positioning, 0.20m in vertical positioning and 0.50 deg. in heading after 10sec of GNSS outage. While this period could tolerate short tunnels, long ones are not considered in the operational design domain (ODD) of the present release of RoboCar. To keep the safety of operation, the precision of pose estimation, e.g. position covariance is monitored and reported to anticipate takeover by the driver. Since the Trimble BX992-INS gives the results of EKF estimated pose in geodesic coordinates, the localization module converts this positioning data into Cartesian  coordinates given reference point/frame, which in this case is the map reference frame. 
In Robocar we use the Earth-North-Up (ENU) convention to get these Cartesian coordinates to represent the vehicle position with respect to the reference point. The localization may be complemented by other techniques to improve reliability in other ODD where the GNSS might prove insufficient \cite{elghazaly2023high, chalvatzaras2022survey, shan2023survey, massa2020lidar} or even dispense of a high precision GNSS altogether. In the scope of RoboCar and given the quality of the onboard GNSS, the localization depends solely on the output of the Trimble. Using map data for localization in GNSS-denied areas is left for our future release of RoboCar.

\subsubsection{Map}

\begin{figure}
    \centering
    \includegraphics[scale=0.156]{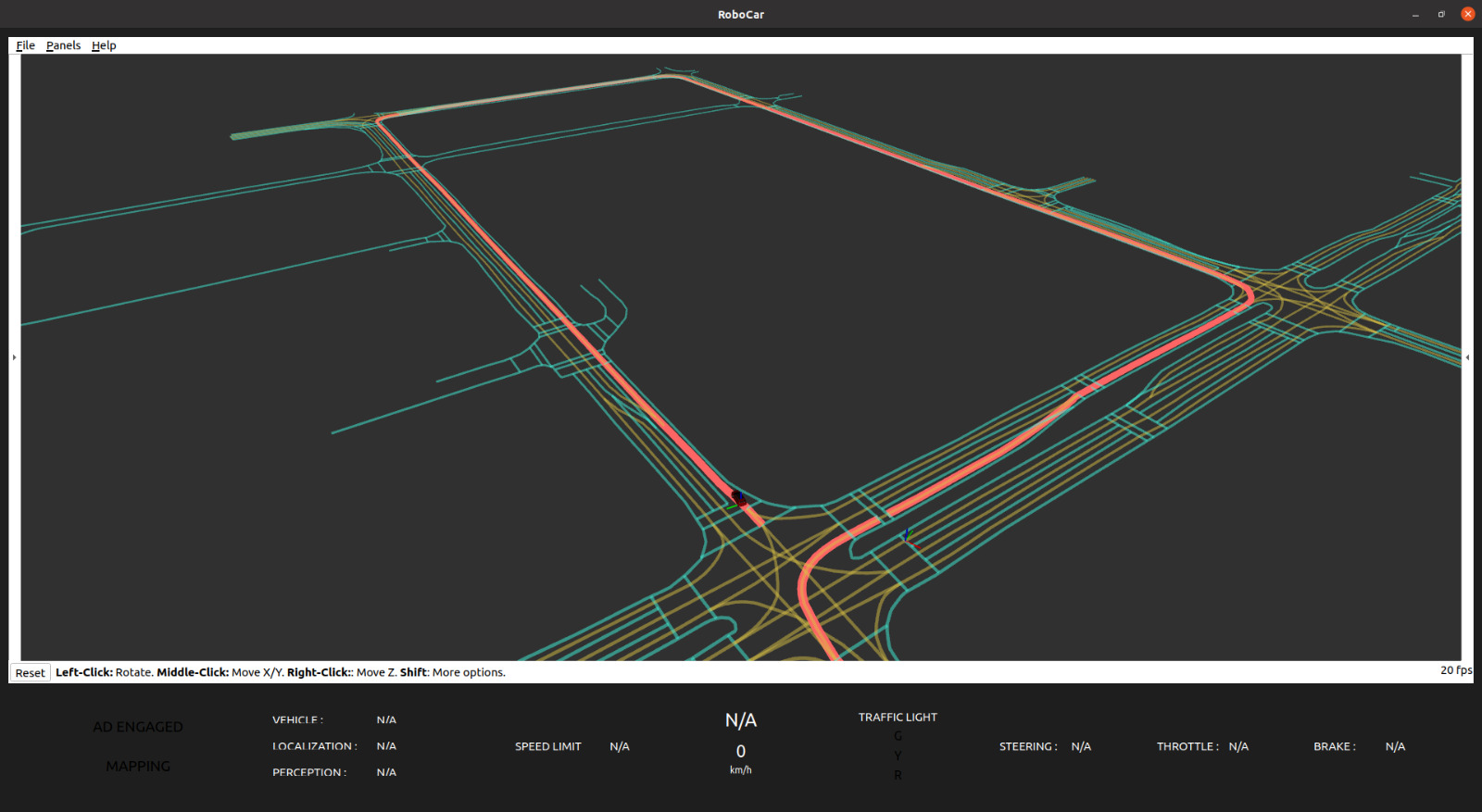}
    \caption{\footnotesize Example of shortest route calculation by the map module of RoboCar using the Belval HD map given the current vehicle position and its final destination point.}
    \label{fig:route}
\end{figure}

This module is responsible for parsing the different layers of an HD map and makes it available to the other core AD modules. An HD map contains lane-level geometric, semantic, and topological information of the road environment, positioned with a few centimeters of precision. \cite{elghazaly2023high}. It is composed of a base layer representing the 3D road environment in the format of a compressed point cloud file (.las/.laz). The geometric, semantic and topological (vector) information of the environment are represented in standard GeoJSON formats \cite{butler2016rfc}. The map module is also in charge of the routing which calculates the shortest path between the current vehicle position and a destination point. 
In RoboCar, the vector layers of the HD maps we provide as open source contain information on how the lane segments are connected. Each lane segment is assigned a unique identifier and the identifiers of the child and parent segments. These geometric and topological information of the map are used to build graph data structures so that calculating the shortest path amounts to a graph search problem for which we could use Dijkstra or $A^{*}$ algorithms as illustrated in Fig.~\ref{fig:route}. The shortest path is then sent to the motion planning module to calculate a collision-free feasible local trajectory. Detailed implementations and other algorithms of the routing component are to be provided in future releases of Robocar.

\subsubsection{Perception}
The perception module contains all components responsible for handling the data coming from perception sensors such as the LiDAR and the cameras. The perception is essential to provide the vehicle with a comprehensive representation of the dynamic environment surrounding it. As such, it can be tasked with detecting objects, lanes, traffic signs, traffic lights, or any other relevant features. In RoboCar, the perception module handles point clouds coming from the LiDAR as well as images coming from the cameras. The LiDAR perception component currently relies on a clustering algorithm, complemented by effective object filtering, and provides the motion planner with a set of localized objects around the vehicle. The camera perception currently handles traffic lights detection and is based on a YOLOv8 model re-trained on the DTLD dataset \cite{8460737}. The model can detect red, yellow, and green lights, and a finite-state machine in the motion planner is used to deal with potential errors and provide filtering. At the time of writing, the perception system does not perform object tracking. However, efforts are in progress to implement a state-of-the-art 3D object perception pipeline based on LiDAR and camera data fusion \cite{deepfusion2022}, which will also include tracking capabilities.

\subsubsection{Planning}
\begin{figure}
    \centering
    \includegraphics[scale=0.18, clip, trim=0.0cm 0.0cm 0.0cm 1.75cm]{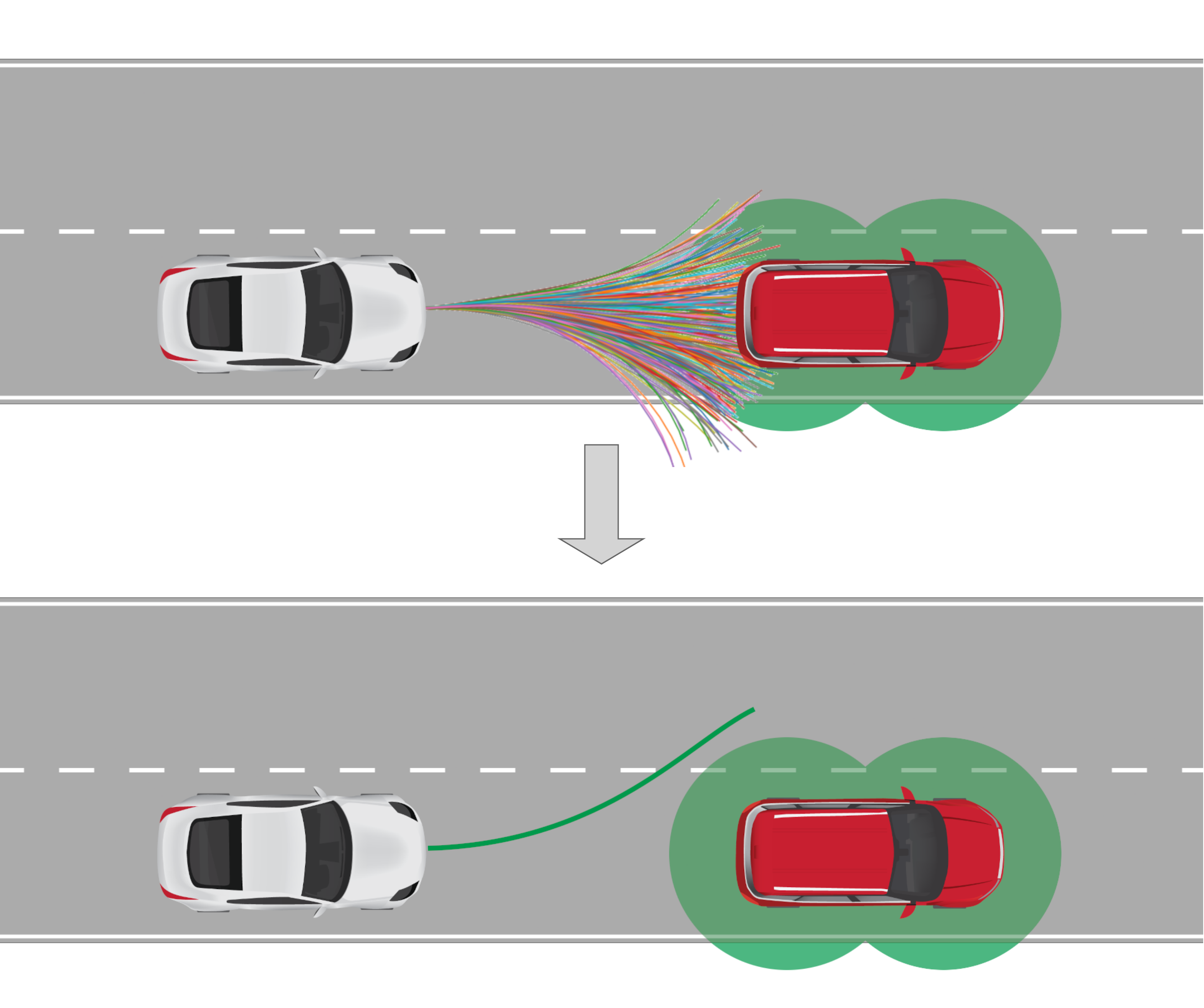}
    \caption{\footnotesize The MPPI-based motion planner generates thousands of trajectories and weighs them using a cost function before merging to obtain the final trajectory. Objects and vehicles are converted into circles for ease of collision calculation.}
    \label{fig:mppi}
\end{figure}

The motion planner is tasked with providing the control with a safe and feasible trajectory to follow under various static constraints imposed by the environment and the dynamic constraints imposed by the dynamic objects surrounding the vehicle \cite{paden2016survey,gonzalez2015review}. Commonly, this problem is solved using either of two approaches. In optimization-based methods, motion planning is formulated as an optimization problem, where the objective is to find the best trajectory satisfying both static and dynamic constraints. On the other hand, the sample-based motion planning methods mainly rely on randomly sampling the configuration space of the vehicle and building a graph or tree structure to explore possible paths \cite{lavalle2006planning}. In RoboCar, the planning module implements a novel approach based on a model predictive path integral (MPPI) algorithm. This MPPI approach uses a sample-based motion planner that generates a great number of trajectories (around 2500) and weighs them using a cost function before merging them to produce a final trajectory as depicted on Fig. \ref{fig:mppi}. The work in the planning module let to \cite{testouri2023towards} which thoroughly explores the implementation of the MPPI planner and evaluates it under real conditions on a closed track. Safe planning also implies respecting a safe distance at all times with other vehicles and objects which are discretized into circles as shown in green on \ref{fig:mppi}. In RoboCar, the safe distance procurement is based on the responsibility-sensitive safety (RSS) model developed by Mobileye \cite{shalevshwartz2018formal} in an attempt to formalize autonomous cars safety. The minimum safe distance $d_{m}$ is computed as follows:

\begin{equation}
d_{m} = \left[ d_0 + v_r\rho + \frac{1}{2} a_{M}\rho^2 + \frac{(v_r + \rho a_{M})^2}{2\beta_{m}} - \frac{v_f^2}{2\beta_{M}} \right]_+
\end{equation}

\noindent where $\left[x\right]_+ := \text{max}(x, 0)$, $d_0$ is an initial offset distance, $\rho$ is the reaction time, $v_r$ is the speed of the rear (ego) vehicle and $v_f$ is the speed of the front vehicle. $a_{M}$ and $\beta_{m}$ are the maximum acceleration and minimum braking of the rear vehicle while $\beta_{M}$ is the maximum braking of the front vehicle. If the actual distance between the front vehicle and the rear vehicle is less than $d_{m}$, the ego vehicle should apply some brake to restore the safe distance.

\subsubsection{Control}
The control is in charge of computing actuation commands to ensure that the vehicle will properly follow the trajectory provided by the motion planner. The control task can be divided into a longitudinal and lateral one. The lateral control implements a Pure Pursuit algorithm \cite{purepursuit} with an adaptive look-ahead distance where the desired steering angle $\delta$ is given by:
\begin{equation}
    \delta = \arctan\left(\frac{2L\sin(\alpha)}{l^t_d}\right),
\end{equation}

\noindent where $L$ is the wheel base and $\alpha$ the heading difference between the target point vector and the vehicle heading. The target point is located on the trajectory at a look-ahead distance $l^t_d$ computed dynamically as follows:
\begin{equation}
l^t_d = \gamma v_t (1 - \kappa_{t}),
\end{equation}

\noindent where $\gamma$ is a fixed parameter, $v_t$ is the current vehicle speed and $\kappa_{t} \in [0,1]$ is a normalized measurement of the curvature of the trajectory. The computed steering angle is then directly sent (after steering ratio conversion) to the actuator as the DriveKit module supports direct steering angle commands. Let $v^r_t$ and $a^r_t$ be the reference speed and acceleration respectively, and let $v^e_t = v^r_t - v_t$ be the measured speed error. The longitudinal control is implemented as follows:

\begin{equation}
\tau_t = K_p a^r_t + K_i \int_{t_0}^t v^e_T dt + K_d \frac{dv^e_t}{dt},
\end{equation}

\noindent where $K_p$, $K_i$ and $K_d$ are fixed parameters and $\tau_t$ is the throttle command. Some refinement should be brought to the longitudinal control, in particular, when the road slope is significant and some limitations became apparent. One could introduce slope coefficients applied to $K_p$, $K_i$ and $K_d$ to remedy the situation.

\subsection{HMI and services}
The human-machine interface (HMI) and services layer implements modules related to visualization and interaction with the autonomous driving software in addition to troubleshooting and software development tools. RoboCar contains three such modules, further described in this section.

\subsubsection{Simulation}
RoboCar implements a simulation module designed to help with the development by offering a virtual test ground for the various algorithms and systems involved. The simulation module contains a component emulating an ego vehicle and components simulating virtual object agents. The vehicle simulation component is based on the kinematic bicycle model\cite{7995816} and on a simple throttle and steering emulation. The development of a more advanced vehicle model is part of future work. Virtual object agents can be implemented by the user to tailor specific behaviors and create adequate driving scenarios.

\subsubsection{Visualization}
\begin{figure}
    \centering
    \includegraphics[scale=0.14]{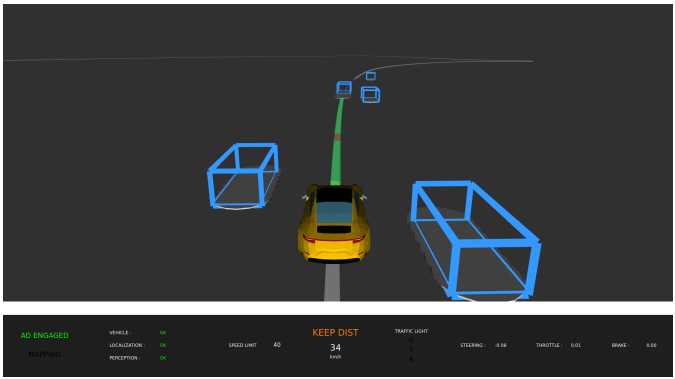}
    \caption{\footnotesize RoboCar running in simulation mode with simulated objects. The center lane is displayed in grey while the MPPI generated trajectory is in green and the target point in red.}
    \label{fig:viz}
\end{figure}

This visualization module contains components allowing the viewing and monitoring of data originating from the vehicle, the sensors and all components of the system such as the perception or the motion planner. It also provides an HMI allowing to engage and disengage the autonomous driving mode including the ability for an operator to apply corrections to the system and even direct inputs to the actuators of the vehicle. The implementation relies on the Qt framework as well as Rviz, Fig. \ref{fig:viz} depicts RoboCar running in simulation mode.

\subsubsection{Logging}
This module extends the ROS2 data recording capabilities. The implemented logging component will automatically record a configurable set of variables when autonomous driving is engaged thus acting as a black box for the system. This data proved useful to troubleshoot the system or fine tune the various algorithms.

\subsection{Main data flows}
\begin{figure}
    \centering
    \includegraphics[scale=0.62, clip, trim=1.0cm 23.0cm 0.0cm 1.0cm]{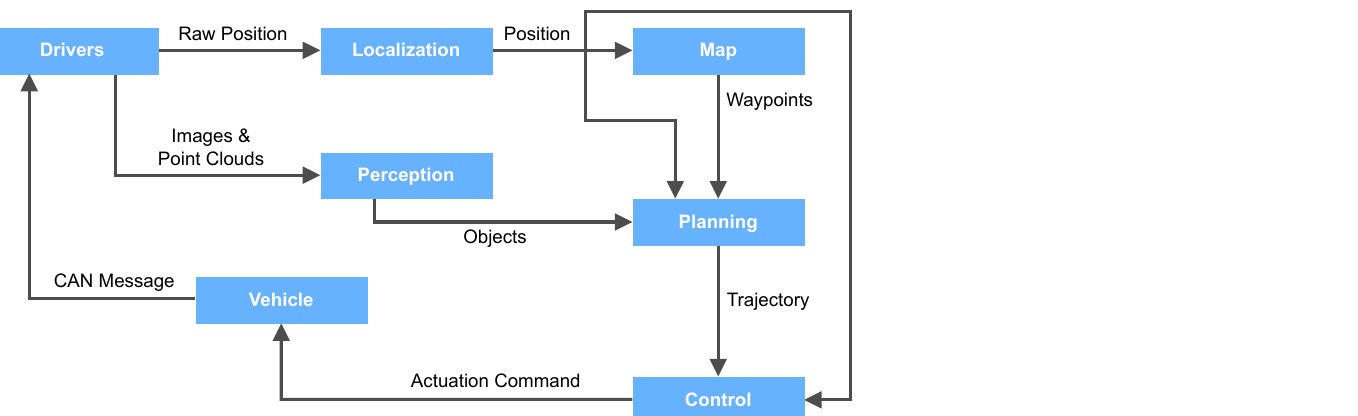}
    \caption{\footnotesize Overview of the main data flows in RoboCar with the ROS2 messages exchanged between the different modules.}
    \label{fig:loop}
\end{figure}

Fig. \ref{fig:loop} depicts an overview of the main data flows in RoboCar. The images and the point clouds are supplied by the sensors drivers to the perception, which in turn provides the motion planner with spatially detected objects. The localization is fed with raw positions coming from the GNSS driver and then produces a referenced position in the map frame. This position is used by the map module to select relevant waypoints according to the routing. In addition to the objects, the position and the waypoints are fed to the motion planner, which computes an optimal and safe trajectory. The control receives both the trajectory and the position and computes optimal actuation commands to properly follow the trajectory. The actuation commands are then sent to the vehicle module which writes the CAN messages to the CAN driver before repeating the entire process.

\section{Experiments}
RoboCar has been thoughtfully tested and validated under a variety of conditions and tasks, with experiments conducted in simulation, on private tracks, and on public roads. This section explores the various experiments undertaken, the preliminary challenges encountered, and presents some quantitative results to provide more insight into the current capabilities of the platform.

\subsection{Authorizations}
To validate our platform under real traffic conditions, we first needed to apply for an experimental permit. The application that has been submitted to the Luxembourg Department of Transport includes two distinct steps. The first was to apply for an authorization for scientific trials, specifying the road network on which the test will be conducted. The second, which was by far the more complex and time-consuming, was to apply for an experimental vehicle registration card. As our base vehicle already has European type approval, we needed to demonstrate that the newly installed hardware components and the software modules that interact with the safety-critical systems of the vehicle are behaving correctly and do not interfere with other control systems. More specifically, we had to demonstrate that in case the autonomous system misbehaves, the safety driver can regain control without compromising the safety of the passengers and other traffic participants. A specialized company was contracted to certify the takeover system, conduct training for the safety driver, and provide guidelines to set up an experimental protocol. After this step, the application could be finalized in order to obtain the new vehicle registration card, which allowed us, together with the authorization for scientific trials, to conduct our experiments on public roads.

\subsection{High-definition mapping}

\begin{figure*}[ht]
    \centering
    \includegraphics[scale=0.50, clip, trim=0.0cm 0.0cm 0.0cm 0cm]{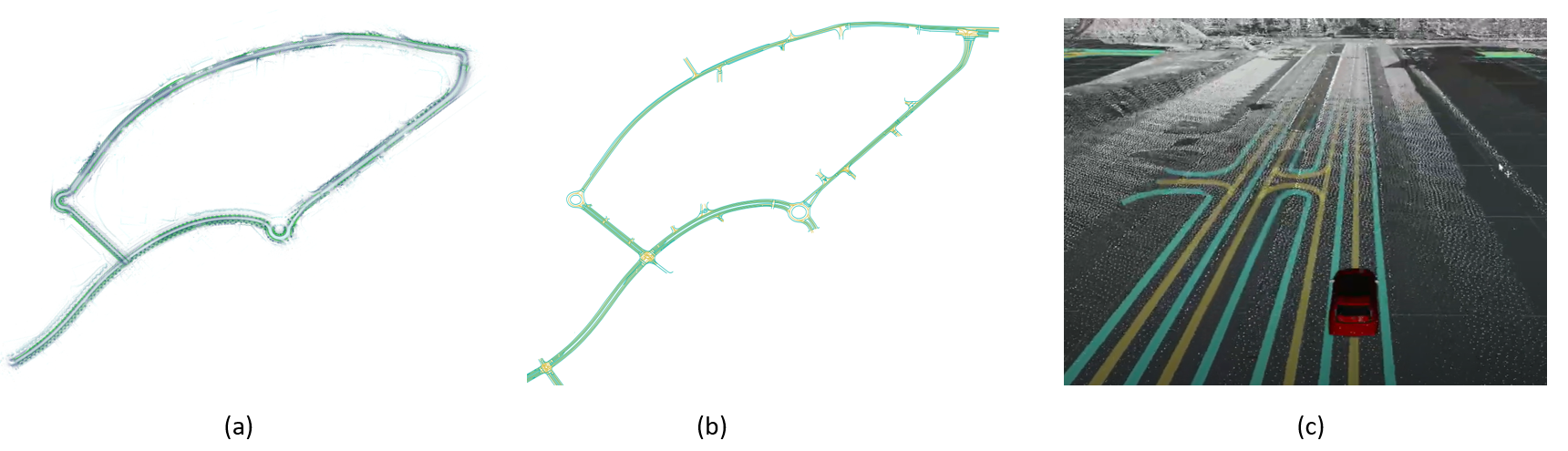}
    \caption{\footnotesize HD map layers of the route used for the public road experiments in Kirchberg. (a) base map layer, (b) geometric and semantic layers (c) all layers superimposed and visualized in RoboCar.}
    \label{fig:hd-map-layers}
\end{figure*}

The high-definition (HD) map is a critical component of autonomous driving systems, providing centimetric and lane-level information needed for localization, perception, scene understanding and motion planning \cite{elghazaly2023high}.

HD maps are composed of several layers, each plays one or more of the above-mentioned roles. The base map layer is usually in the form of a point cloud representing the 3D details of the environment. The geometric layer provides a vectorized lane-level representation of road geometry using basic geometric primitives (points, line segments, and polygons). While the semantic layer assigns semantic features to these primitives, the road connectivity layer defines how these geometric primitives are connected. 
In the current release of RoboCar, we provide two examples of HD maps that we have created for our public road testing. The first is a public road track of about 3km in Kirchberg (Fig.~\ref{fig:hd-map-layers}) while the second is a public road track of about 2km in Belval (Fig.~\ref{fig:route}). In RoboCar, we create the base map layer (Fig.~\ref{fig:hd-map-layers}.a) using an adapted version of the FAST-LIO mapping algorithm \cite{xu2021fast} that robustly fuses the LiDAR point cloud of the Ouster OS1 and INS data of the Trimble BX922 using a tightly-coupled iterated Kalman filter. The original FAST-LIO fuses the LiDAR point cloud and IMU to create a point cloud map while simultaneously estimating vehicle odometry and positional data. We take advantage of the high-precision of INS data of the Trimble BX922 positional data as an additional input in the estimation of the map to avoid the well-known loop closure problem. This mapping process generates a point cloud base map in Cartesian coordinates which is then converted to geodesic coordinates via georeferencing. 
The base map is imported into the geographical information system (GIS) software QGIS\footnote{https://qgis.org}. 
The geometric, semantic and road connectivity layers are then created. 
First, geometric primitives of the geometric layer are manually crafted by matching and identifying features in the point cloud map’s intensity and colors.
Next, these elements are traced to assign semantic attributes and define their connections. Finally, the vectorized layers (Fig.~\ref{fig:hd-map-layers}.b) are exported to a GeoJson format for easy use in RoboCar. All layers are superimposed altogether in Fig.~\ref{fig:hd-map-layers}.c.

\subsection{Early experiments}
Early experiments involved simulations (see Fig. \ref{fig:viz}) and real world tests on a small test track around the campus of the University, which allowed us to validate the initial version of the system. Those experiments also proved useful to validate new algorithms and check potential system regressions throughout the continued development cycle. Once RoboCar was deemed functional enough and prior to public road experiments, it underwent extensive testing on a closed training track outside Luxembourg City. This allowed to safely assess and calibrate RoboCar in a more realistic setting and proved to be a vital step towards hitting the public road. An example of an experimental setup conducted on the track can be found in \cite{testouri2023towards} where the MPPI based motion planner was tested and evaluated.

\subsection{Public road experiments}
\begin{figure*}
    \centering
    \includegraphics[width=\textwidth]{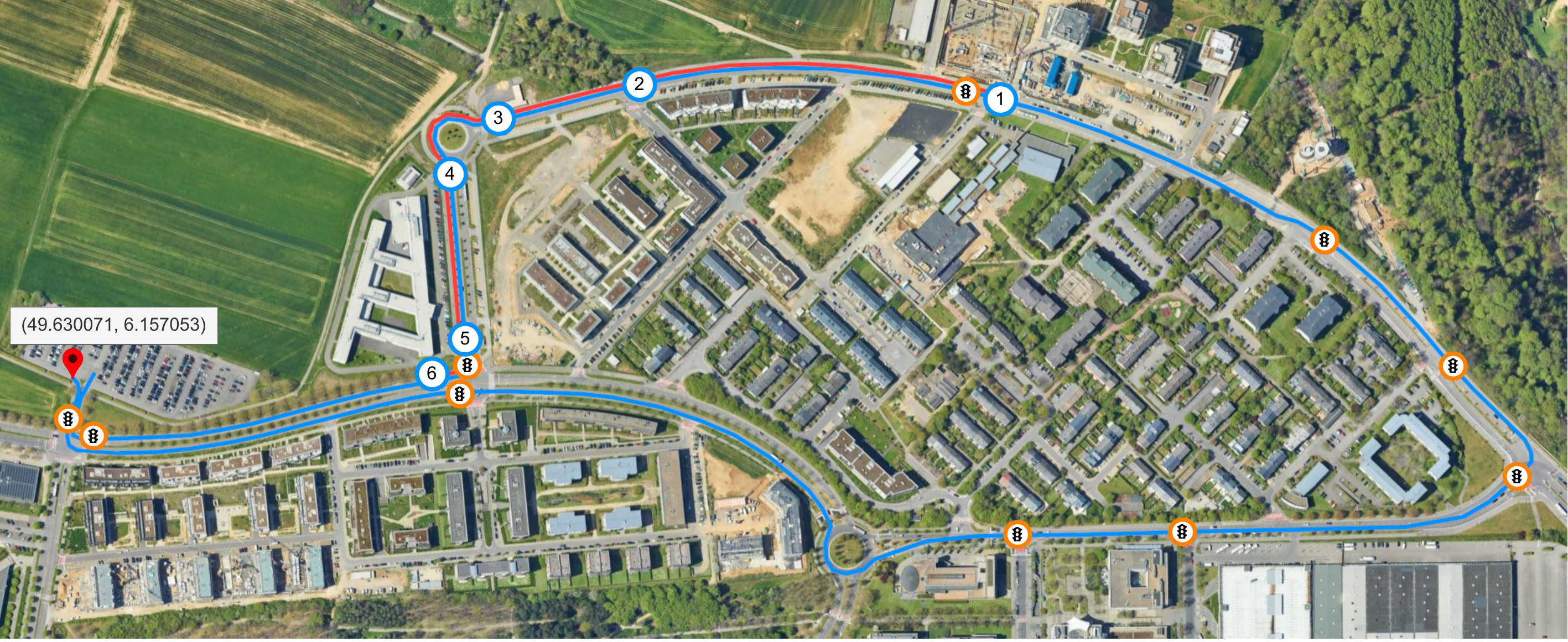}
    \caption{\footnotesize Route used for the public road experiments on Kirchberg. The loop provides various driving scenarios while remaining accessible for early prototypes. Traffic lights encountered are positioned on the map with their corresponding markers. The red portion delimits the section of the path illustrated on Fig. \ref{fig:results} along with the corresponding front camera snapshots numbers.}
    \label{fig:kirchberg_path}
\end{figure*}

Various public road experiments and demonstrations were conducted, most notably a public demonstration was performed to showcase the capabilities of the platform and raise awareness about autonomous driving technologies to a wider audience. The public road experiments were centered around an approximate 3km loop in the Kirchberg area of Luxembourg City as shown on Fig. \ref{fig:kirchberg_path}. This path was chosen because it provided a diverse set of driving scenarios under moderate traffic conditions while remaining accessible for experimental ADS.

\subsection{Experimental setup}
In the context of this paper, a dedicated road experiment was conducted on the same route previously described to evaluate the latest capabilities of RoboCar. The speed limit $v_{max}$ was set to 40 km/h and leaving the center lane was not allowed, meaning that the vehicle would have to stop on every blocking obstacle encountered. The traffic light perception relied on the Intel RealSense camera placed inside the cabin to validate the system in its simplest and most cost-effective configuration. The information on which traffic light to follow along the path was known in advance and encoded in the map so that the motion planner can select the proper traffic light from the perception. The parameters of the RSS used for the experiment are summarized on Table \ref{tab:rss_params}.

\begin{table}[!t]
    \begin{center}
        \caption{\scshape Public road experiment parameters.}
    \begin{tabular}{p{3cm} | p{1cm} | p{1cm}}
        \hline
        \hline
        Parameter & Value & Unit\\
        \hline
        \hline
        $v_{max}$ & 40 & $km/h$\\
        $d_0$ & 7.0 & $m$\\
        $\rho$ & 0.3 & $s$\\
        $a_{max}$ & 2.5 & $m/s^2$\\
        $\beta_{min}$ & 1.5 & $m/s^2$\\
        $\beta_{max}$ & 9.0 & $m/s^2$\\
        \hline
        \hline
    \end{tabular}
    \label{tab:rss_params}
    \end{center}
\end{table}

\subsection{Results}
\begin{figure*}[ht]
    \centering
    \includegraphics[width=\textwidth]{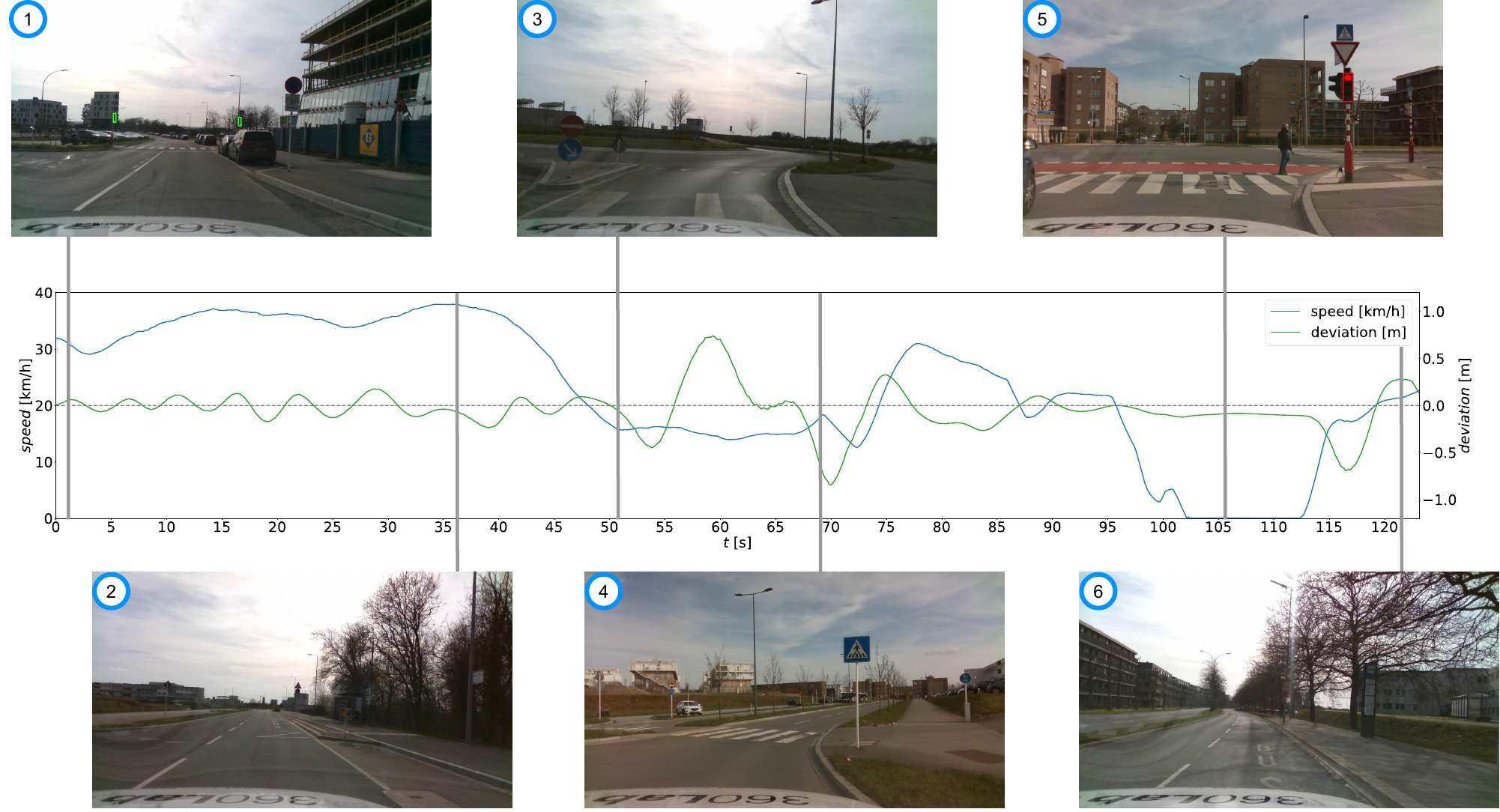}
    \caption{\footnotesize Speed and center lane deviation along a section of the Kirchberg route used for the experiments. The speed transitions are smooth and the speed limit is never exceeded. The traffic light detection is both timely and accurate, facilitating prompt responses from the vehicle. The deviation from the center lane is small expect in the roundabout and tight turns where it becomes more significant.}
    \label{fig:results}
\end{figure*}

The experimental vehicle running RoboCar successfully completed the Kirchberg loop while respecting the speed limits and traffic lights. A safe distance from a leading vehicle was also properly maintained, requiring no manual intervention. However, there is room for improvement in the smoothness of the car-following motion, as the vehicle tended to oscillate around the safe distance. On one occasion, the intervention of the safety driver was required to correct a slight deviation from the path. The traffic lights were on average detected by the system at a distance of $44m$, which proved sufficient to smoothly stop the vehicle once the signal turned red. Upon investigation, the limiting factors of the detection distance seemed to be the map module, which would not always load the upcoming traffic lights in a timely manner as well as the curvature of the road combined with obstacle obstructing the sight. The perception system in itself was able to detect traffic lights from a distance of up to $70m$. Fig. \ref{fig:results} illustrates a snippet of the results along a section of the Kirchberg route used for the experiments. The speed of the vehicle and an estimate of the center lane deviation are reported alongside snapshots of the front camera. Snapshots 1 and 2 show the vehicle in a cruising phase properly detecting a green traffic light at a distance and observing the speed limit. Some oscillations around the center lane can be observed, however, their amplitude and period make them barely perceptible for a passenger onboard. Snapshot 3 illustrates the vehicle entering the roundabout having significantly reduced its speed. It can be observed throughout the roundabout that the center lane deviation can get more significant at about $0.75m$. Snapshot 4 is taken right before one manual intervention that was required at around $t=69s$ to avoid the car drifting too far away from the center lane, risking grazing the sidewalk. This center lane deviation could be the result of inaccuracies produced by the motion planner and/or the control module. The autonomous driving was re-engaged a few seconds later around $t=72$. Snapshot 5 shows the car fully stopped at a red light while snapshot 6 is taken after the green light with the vehicle being on a smooth acceleration path to another cruising phase. The right turn following the traffic light on snapshot 5 being quite tight, it can again be observed that the deviation was more significant. Overall, the speed transitions were smooth and allowed for a comfortable driving.

\section{Discussion}
This section discusses the pragmatic scope of RoboCar and derives relevant limitations and future works based on the experiments presented in the previous section.

\subsection{Platform scope}
The experiments demonstrated the ability of RoboCar and its hardware platform to enable basic self-driving operations on public roads, however, some limitations remain requiring additional work to reach full autonomy reliably. The hardware platform itself has limitations and also constraints further development due to a lack of sufficient computing power and limited sensors among other things. In this context, RoboCar is not positioned as full-blown replacement of existing autonomous driving solutions targeting production deployments but rather as a foundational framework allowing researchers to develop and test new techniques and algorithms related to connected and autonomous mobility. Compared to more advanced and complex platforms such as Apollo and Autoware, RoboCar is a lightweight, easy-to-deploy autonomous driving stack designed for accessibility and scalability. While Apollo and Autoware offer extensive features, they come with significant complexity, numerous external dependencies, and are not well-suited for the European regulatory and operational landscape. RoboCar, in contrast, has a low footprint and can seamlessly scale from a small RC car to a full-blown autonomous vehicle. Its hardware requirements are minimal, making it compatible with a wide range of inexpensive computing platforms. Unlike Apollo, RoboCar is built on ROS2, enhancing modularity and interoperability with other projects. By focusing on essential functionality without unnecessary overhead, RoboCar offers a practical, efficient, and accessible solution for autonomous driving development.

\subsection{Limitations and future works}
The aim of the RoboCar initiative is not to create a fail-safe autonomous driving system, but rather to establish an experimental platform for researchers and innovators. It allows for meeting a smaller set of requirements in terms of autonomous driving capabilities and safety. Therefore, it is important to highlight the main limitations within the context of a research platform for autonomous driving. 

The current main area for improvement is within the perception module. Enhancements are needed to increase the accuracy of the 3D object detection, reduce the number of false positives, and incorporate robust object tracking and prediction capabilities. An accurate deep learning 3D object detection solution based on a fusion between camera and LiDAR data that also provides robust tracking and prediction is currently being evaluated and should be released in due time.

Inaccuracies in both lateral and longitudinal planning and control persist, primarily as deviations from the center lane during tighter turns and non-smooth car-following motion. Therefore, further adjustments and tuning are needed to address these issues.

A minor issue involves improving the simulation model of the ego vehicle to help with the overall system development. Finally, the hardware platform itself is also limited in computing power and sensors, future works could also explore the possibility of upgrading and or extending the hardware set while keeping the overall platform relatively inexpensive.

While RoboCar is designed to be lightweight and cost-effective, scalability and complexity are not limitations but opportunities for growth. Our primary goal is to foster engagement within the European research community, encouraging collaboration through open-source contributions, research partnerships, and potentially EU-funded projects. By providing a solid, easy-to-use foundation, we aim to empower researchers and developers to extend RoboCar for more complex use cases, from urban autonomy to large-scale fleet deployments. Rather than building a monolithic system upfront, we believe in a modular, community-driven evolution, ensuring that RoboCar remains adaptable, interoperable, and well-suited for the unique challenges of autonomous mobility in Europe.

\section{Conclusion}
In this paper, we presented RoboCar, a ROS2-based modular and easy-to-deploy autonomous driving software written in C++ and Python, targeted for research and prototyping purposes. The architecture of RoboCar and its modules are detailed alongside a hardware platform focused on cost efficiency, providing a comparatively inexpensive yet capable autonomous driving vehicle. Among RoboCar's innovative features is an MPPI-based motion planner that also utilizes the RSS standard for safe distance calculation. RoboCar was tested on public roads and demonstrated its ability to perform initial self-driving operations by successfully completing a loop through various driving situations in public traffic in the Kirchberg area of Luxembourg City. Overall, the system performed well, mostly operating autonomously within the traffic flow. Finally, we discussed the scope and limitations of the platform compared to state-of-the-art systems. RoboCar is available to anyone at \href{https://github.com/sntubix/robocar}{\textcolor{magenta}{https://github.com/sntubix/robocar}} and is released under an open-source MIT license.

\section*{Acknowledgments}
The authors would like to thank all partners that supported this initiative over the years.

\bibliographystyle{IEEEtranS}
\bibliography{bibliography}

\begin{IEEEbiography}
[{\includegraphics[width=1in,height=1.25in,clip,keepaspectratio]{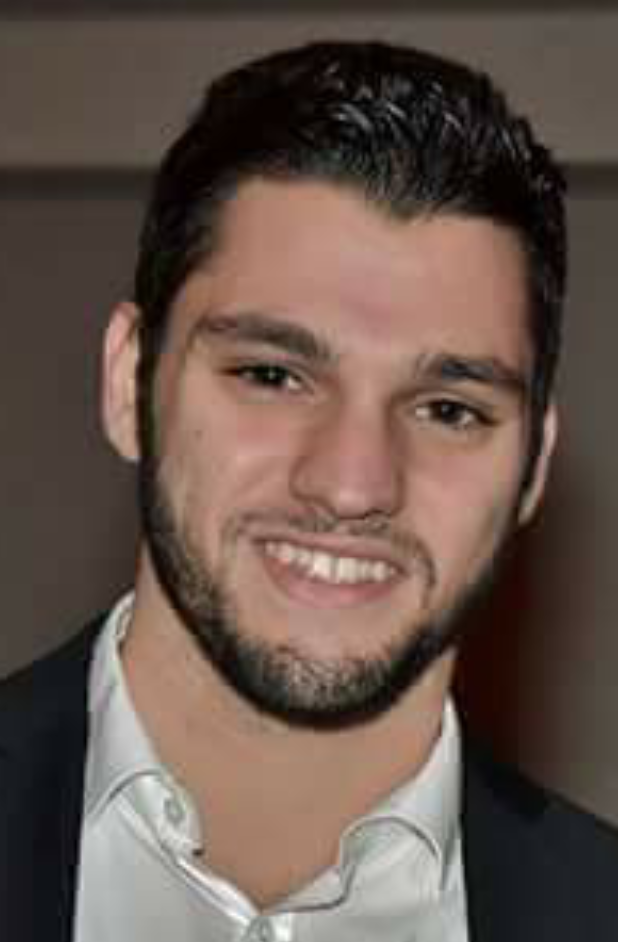}}]{MEHDI TESTOURI} received his Master's Degree in Computer Science and Engineering from the University of Liège in 2020. In 2021, he joined the Interdisciplinary Centre for Security, Reliability and Trust (SnT) of the University of Luxembourg where he worked at the 360Lab on various autonomous driving research projects. In 2023, he joined the UBIX Research Group at SnT, focusing on ubiquitous and intelligent systems, as a Research and Development Specialist. His research interests are mainly autonomous driving systems, computer vision and deep learning. Email: mehdi.testouri@uni.lu.
\end{IEEEbiography}

\begin{IEEEbiography}[{\includegraphics[width=1in,height=1.25in,clip,keepaspectratio]{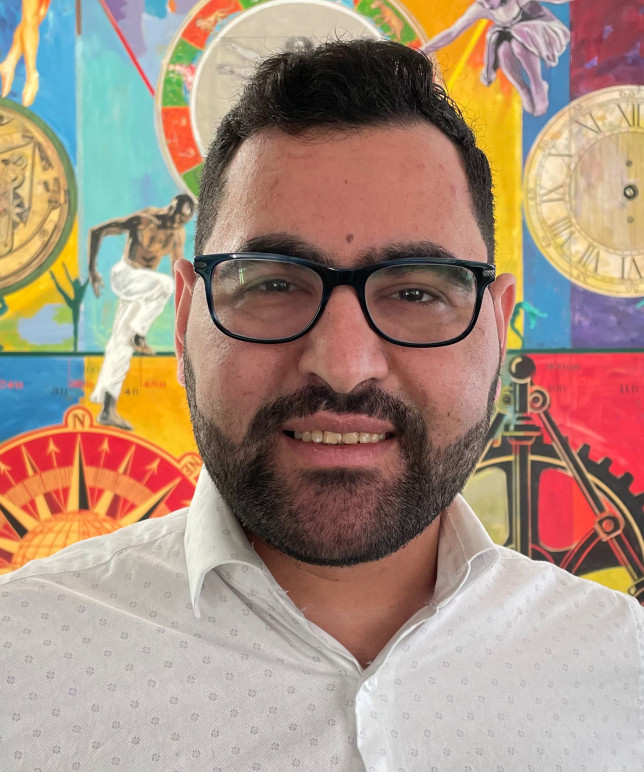}}]{GAMAL ELGHAZALY} (IEEE Member) 
received his PhD in Robotics from the University of Montpellier, France in 2017, where he conducted research at LIRMM-CNRS in modelling, motion planning and control for innovative robotics. He received the MSc Degree in Robotics in 2013 jointly from both Ecole Centrale de Nantes, France and the University of Genova, Italy under the framework of the European Master on Advanced Robotics (EMARO).
Currently, Dr. Elghazaly is a Postdoctoral Researcher at the Interdisciplinary Centre for Security, Reliability and Trust (SnT) of the University of Luxembourg.
He is a member of the UBIX Research Group at SnT, focusing on ubiquitous and intelligent systems. His research expertise is mainly focused on safe and scalable autonomous driving. 
He is also co-heading the 360Lab of SnT to push the limits of these research topics. 
He joined Milla Group in 2017, where he was a senior R\&D Engineer, leading the software development and R\&D activities. Later on, he has been promoted as Chief Scientific Officer of Milla, where he was in charge of defining and implementing strategic research plans to keep pace with such a rapidly growing technology. Email: gamal.elghazaly@uni.lu.
\end{IEEEbiography}

\begin{IEEEbiography}[{\includegraphics[width=1in,height=1.25in,clip,keepaspectratio]{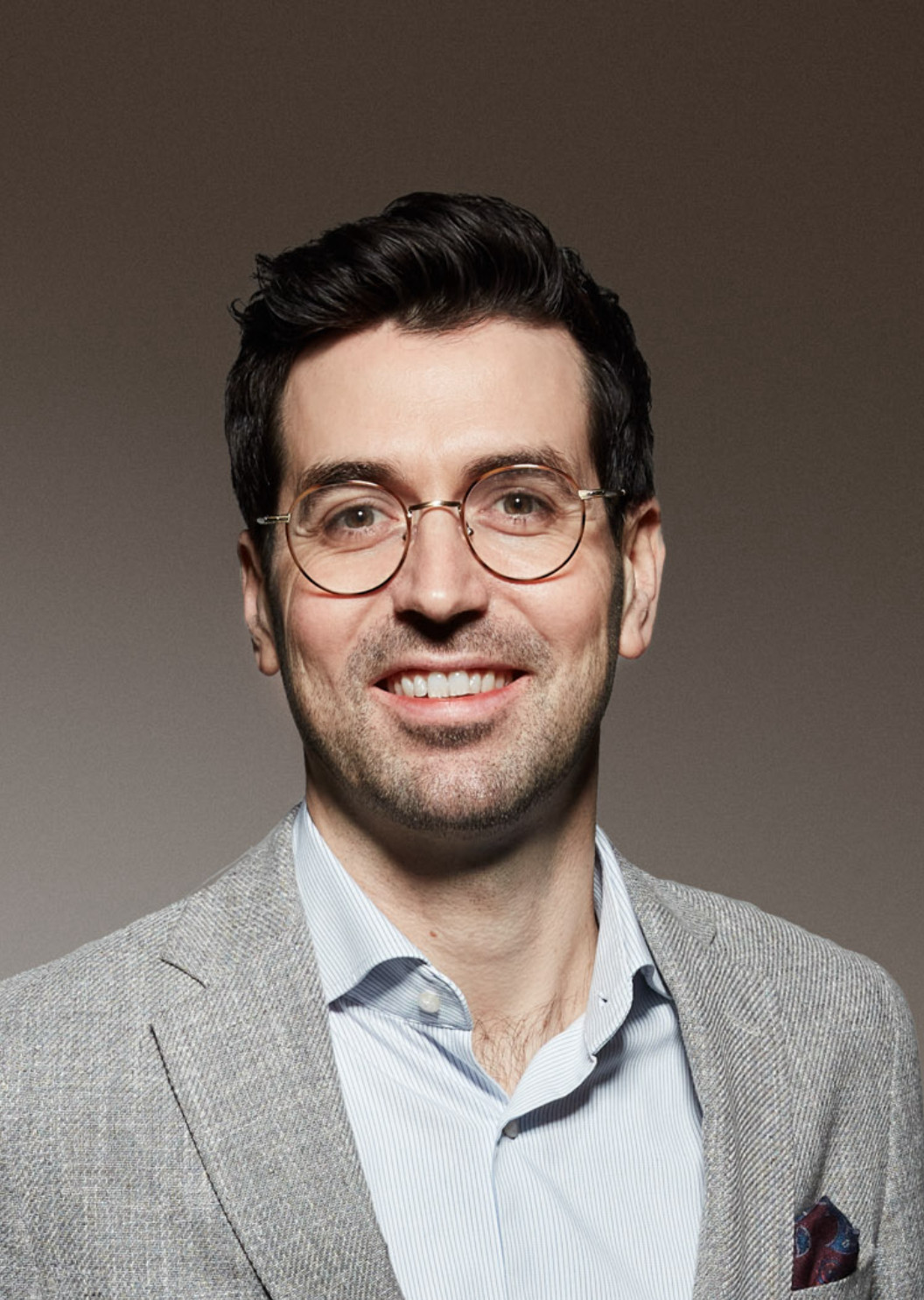}}]{RAPHAEL FRANK } (Senior IEEE Member) received his Ph.D. in Computer Science from the University of Luxembourg in 2010. During his Ph.D. studies, he was a visiting scholar at the University of California, Los Angeles (UCLA), where he conducted research on data routing protocols for vehicular networks. In 2006, he received his Master's Degree in Cryptography and Network Security from the University Joseph Fourier in Grenoble, France. He is currently an Assistant Professor at the Interdisciplinary Centre for Security, Reliability and Trust (SnT) of the University of Luxembourg where he heads the UBIX Research Group. His research expertise covers topics in the areas of distributed and communicative systems, machine learning, and IoT. He is habilitated to supervise Ph.D. students and lead large research projects as Principal Investigator (PI). Email: raphael.frank@uni.lu.
\end{IEEEbiography}

\end{document}